\documentclass{article} 
\usepackage{iclr2023_conference,times}

\usepackage{amsmath,amsfonts,bm}

\def\eqref#1{equation~\ref{#1}}

\def\1{\bm{1}}

\DeclareMathAlphabet{\mathsfit}{\encodingdefault}{\sfdefault}{m}{sl}
\SetMathAlphabet{\mathsfit}{bold}{\encodingdefault}{\sfdefault}{bx}{n}

\def\gL{{\mathcal{L}}}

\newcommand{\E}{\mathbb{E}}

\usepackage{hyperref}
\usepackage{url}

\title{DP-Adam: Correcting DP Bias in Adam's Second Moment Estimation}

\author{Qiaoyue Tang, Mathias L\'ecuyer \\
Department of Computer Science\\
University of British Columbia\\
Vancouver, Canada \\
\texttt{qiaoyuet@cs.ubc.ca, mathias.lecuyer@ubc.ca}
}

\usepackage{mathtools}
\usepackage{caption}
\usepackage{subcaption}
\usepackage{graphicx}
\usepackage{multirow}
\usepackage{pifont}
\usepackage[ruled,vlined]{algorithm2e}
\SetKwInOut{Hyperparameter}{Hyperparameter}
\usepackage{amsmath}
\usepackage{amssymb}
\usepackage[english]{babel}

\usepackage{enumerate}
\usepackage{enumitem}
\usepackage{tikz}

\def\cN{\mathcal{N}}
\def\I{\mathbb{I}}

\def\V{\mathbb{V}}

\iclrfinalcopy 
\begin{document}

\maketitle

\begin{abstract}
We observe that the traditional use of DP with the Adam optimizer introduces a bias in the second moment estimation, due to the addition of independent noise in the gradient computation.
This bias leads to a different scaling for low variance parameter updates, that is inconsistent with the behavior of non-private Adam, and Adam's sign descent interpretation.
Empirically, correcting the bias introduced by DP noise significantly improves the optimization performance of DP-Adam.
\end{abstract}

\section{Introduction}
The Adam optimization algorithm \citep{orig_adam} is the default optimizer for several deep learning architectures and tasks, most notably in Natural Language Processing (NLP) and on graph data.
Even in vision tasks where Adam is less prevalent, it typically requires less parameter tuning than SGD to reach good performance.
However we observe that when combined with Differential Privacy (DP), Adam does not perform as well: it suffers a large degradation of performance compared to SGD on vision tasks, and NLP and graph tasks perform poorly when training from scratch.
Recent empirical investigations suggest that Adam's performance stem from its update rule performing a smooth version of sign descent \citep{kunstner2023heavytailed}. The key components of Adam's update are two exponential moving averages estimating the first and second moments of mini-batch gradients.
We show that while DP noise does not affect the first moment, it does add a constant bias to the second.
This additive change to the second moment moves the Adam update away from that of sign descent, by scaling gradient dimensions based on their magnitude.
We show that we can correct for this DP noise induced bias.
Empirically, correcting Adam's second moment estimate for DP noise significantly increases test performance for Adam with Differential Privacy, on both vision and NLP tasks.

\section{The Adam update under Differential Privacy}
\label{sec:method}
The Adam update \citep{orig_adam} is defined as follows, let $g_t = (1/B)\nabla f(\theta_{t-1})$ be the average gradient over a mini-batch of size B with respect to loss function $f$ at step $t$; let $\beta_1$ and $\beta_2$ be Adam's decay coefficients: at each step, Adam updates $m_{t} \leftarrow \beta_{1} m_{t-1} + \left(1-\beta_{1}\right) g_t$, \, $\widehat{m}_{t} \leftarrow m_{t} /\left(1-\beta_{1}^{t}\right)$ and $v_{t} \leftarrow \beta_{2} v_{t-1}+\left(1-\beta_{2}\right) g_t^{2}$, \, $\widehat{v}_{t} \leftarrow v_{t} /\left(1-\beta_{2}^{t}\right)$.
The model's parameters are updated as $\theta_t \leftarrow \theta_{t-1} - \eta \Delta_t$, with learning rate $\eta$, $\Delta_t = \hat{m}_t / (\sqrt{\hat{v}_t} + \gamma)$ the update direction, and $\gamma>0$ a small constant added for numerical stability. 
Intuitively, Adam's $m_t$ and $v_t$ use an exponential moving average to estimate $\E[g_t]$ and $\E[g_t^2]$, the vector of first and second moment of each parameter's gradient, respectively.

Recent evidence \citep{kunstner2023heavytailed} supports the hypothesis that Adam derives its empirical performance from being a smoothed out version of sign descent. At a high level, Adam performs well in settings (e.g., NLP) where sign descent also performs well, at least when running with full (or very large) batch.
We next describe Adam's update rule under this sign descent hypothesis, before working out the impact of DP noise on this interpretation:
\begin{enumerate}
\item If for parameter $i$, $|\E{[g_t]}|_i \gg \sqrt{\V{[g_t]}_i}$, then the update's direction is clear. And since $|\E{[g_t]}|_i \approx \sqrt{\E{[g^2_t]}_i}$, the Adam update is $\E{[g_t]}_i / \sqrt{\E{[g^2_t]}_i} \approx \pm 1$, and Adam is sign descent. Updates are {\em not scaled based on $|\E{[g_t]}|_i$}.
\item If for parameter $i$, $|\E{[g_t]}|_i \not\gg \sqrt{\V{[g_t]}_i}$, the sign is less clear and Adam's update is in $[-1, 1]$, scaled closer to $0$ the more uncertain the sign is (smoothing behavior).
\end{enumerate}
Finally, Adam ensures numerical stability when $|\E{[g_t]}|_i \approx 0$ and $\V{[g_t]}_i \approx 0$ using the additive constant $\gamma$ in the denominator of the update. In that case, the update is approximately $\E{[g_t]}_i/\gamma \approx 0$.

To summarize, under the sign descent hypothesis, Adam  updates parameters with low variance gradients with a constant size $\pm 1$ update (or $\pm \eta$ after the learning rate is applied), and rescales the update of parameters with high variance gradients towards $0$. As we describe next, adding Differential Privacy to gradient computations breaks this interpretation of Adam as sign descent.

\paragraph{Using Adam with Differential Privacy.}
Most optimization approaches for deep learning models with Differential Privacy (DP) follow a common recipe: compute each update (averaged gradients over a mini-batch) with DP, and leverage DP's post-processing guarantee and composition properties to analyse the whole training procedure.
Computing a DP update over a mini-batch involves clipping per-example gradients to control the update's sensitivity, and adding {\em independent} Gaussian noise to the aggregated gradients. Formally: for each step $t$, let $g_{n} = \nabla f(\theta_t, x_n)$ be the gradient for sample $n$, and let $C$, $\sigma$ be the maximum $L2$-norm clipping value and the noise multiplier, respectively. For a mini-batch $B$, $\overline{g}_t = (1/B)\sum_{n} g_{n} / {\max{(1, \lVert g_n \rVert_{2}/C})}$ is the mean of clipped gradients over the minibatch, which is a biased estimate of $g_t$. Then, the DP gradient update is $\Tilde{g}_t = \overline{g}_t + (1/B) z_t, \; z_t \sim \cN(0, \sigma^{2}C^{2}\I^{d})$.
With this recipe, any optimizer that only takes mini-batch updates as input, such as Adam, can be applied to the DP update $\Tilde{g}$ and preserve privacy. This is how existing DP approaches using Adam work (e.g., \cite{li2021}), yielding the following update: let the superscript $p$ denote private version of a quantity, then $m_{t}^{p} \leftarrow \beta_{1} m_{t-1}^{p} + \left(1-\beta_{1}\right) \Tilde{g}_t$, \, $\widehat{m}_{t}^{p} \leftarrow m_{t}^{p} /\left(1-\beta_{1}^{t}\right)$, \, $v_{t}^{p} \leftarrow \beta_{2} v_{t-1}^{p}+\left(1-\beta_{2}\right) \Tilde{g}_t^{2}$, \, $\widehat{v}_{t}^{p} \leftarrow v_{t} /\left(1-\beta_{2}^{t}\right)$, $\theta_t \leftarrow \theta_{t-1} - \eta \hat{m}_t / (\sqrt{\hat{v}_t} + \gamma)$.

\paragraph{DP noise biases second moment estimates, breaking the sign descent behavior.}
Under DP, Adam estimates the first and second moments as $m_t^{p}$ and $v_t^{p}$, using $\Tilde{g}_t$ in order to preserve privacy. Since the noise added for DP is independent of the gradient update, there is no impact on the first moment estimate in expectation:
\begin{align}
\label{eq:mt}
    \E{[m_t^{p}]} = \E{\bigg[ (1-\beta_{1}) \sum_{\tau=1}^{t} \beta_{1}^{t-\tau} \Tilde{g}_\tau \bigg]} = (1-\beta_{1}) \sum_{\tau=1}^{t} \beta_{1}^{t-\tau} \bigg( \E{[ \overline{g}_\tau ]} +  \underbrace{\frac{1}{B}\E{[z_\tau]}}_{\text{0}} \bigg) = \E{[m_t^{c}]}.
\end{align}
However, $v_t^{p}$ is now a biased estimate of the second moment of the mini-batch's update $\bar{g}_t$, as it incurs a constant shift due to DP noise. By independence of the DP noise $z_t$ and $\bar{g}_t$, we have that:

\begin{align}
\label{eq:vt}
    \E{[v_t^{p}]} = \E{\bigg[ (1-\beta_{2}) \sum_{\tau=1}^{t} \beta_{2}^{t-\tau} {\Tilde{g}_\tau}^{2} \bigg]} = \underbrace{(1-\beta_{2}) \sum_{\tau=1}^{t} \beta_{2}^{t-\tau} \E{ [ \overline{g}_\tau^{2}]}}_{\E{[v_t^{c}]}} + \underbrace{(1-\beta_{2}^{t})\bigg(\frac{\sigma C}{B} \bigg)^{2}}_{\Phi}, 
\end{align}
where $\E{[m_t^{c}]}$ and $\E{[v_t^{c}]}$ are the quantities estimated under regular Adam, computed with respect to $\bar{g}_t$ (due to DP clipping). Under exact expectations, the Adam update becomes $\Delta_t = 
\E{[\tilde{g}_t]} / \sqrt{\E{[\tilde{g}_t^2]}} = \E{[\bar{g}_t]} / \sqrt{\V{[\bar{g}_t]+\E(\bar{g}_t)^2}+\Phi}$.

To understand the implication of DP noise bias $\Phi$, let us follow \cite{orig_adam} and interpret the update under the assumption that $\E{[m_t^{c}]} \approx (1-\beta_1^{t})\E{[\bar{g}_t]}$ and $\E{[v_t^{c}]} \approx (1-\beta_2^{t})\E{[\bar{g}_t^{2}]}$---e.g., the first and second moment $\E{[\bar{g}_\tau]}, \E{[\bar{g}_\tau^{2}]}, \tau=0, 1, 2, \ldots, t$ are stationary; or decay coefficients $\beta_1, \beta_2$ are such that past $g_\tau$ are assigned small weights,  $t$ is large enough, and $\beta_1, \beta_2$ are such that $(1-\beta_1^{t})/{\sqrt{1-\beta_2^{t}}} = 1$.
Let $\Phi$ be the bias accumulated from DP noise variance up to step $t$ in $v_t^{p}$ (Eq. (\ref{eq:vt})). This bias in the second moment rescales the Adam update, which becomes incompatible with the sign descent interpretation. Focusing on the sign descent regime---when a parameter $i$ in the model has a large signal and small variance, such that $|\E{[\bar{g}_t]}|_i \approx \sqrt{\E{[\bar{g}^2_t]}_i}$---the Adam update becomes $\pm (|\E{[\bar{g}_t]}|_i / \sqrt{\E{[\bar{g}^2_t]}_{i} + \Phi})$ instead of $\pm 1$.
For example: if $|\E[\bar{g}_t]|_i=\sqrt{0.1\Phi}$, the update will be $\approx \pm 0.1$, whereas it will be $\approx \pm 1$ if $|\E[\bar{g}_t]|_i=\sqrt{10\Phi}$. In each case, without DP noise Adam would result in a $\pm 1$ update.

Importantly, re-scaling the learning rate $\eta$ is not sufficient to correct for this effect. Indeed, consider two parameters of the model indexed by $i$ and $j$ that, at step $t$, both have updates of small variance but different magnitude, say $|\E[\bar{g}_t]|_i=\sqrt{0.1\Phi}$ and $|\E[\bar{g}_t]|_j=\sqrt{10\Phi}$.  Then the Adam update for $i$ will be $\approx \pm 0.1$ and that of $j$ $\approx \pm 1$, and no uniform learning rate change can enforce a behavior close to sign descent for both $i$ and $j$ in this step.

\paragraph{Correcting for DP noise in DP-Adam.} 
Since we can compute the bias in $v_t^{p}$ due to DP noise (see Eq. (\ref{eq:vt})), we propose to correct for this bias by changing the Adam update $\Delta_t$ as follows:

\begin{equation}
\label{eq:adam_corr}
    \Delta_t = \hat{m}_t / \sqrt{\max{\big(\hat{v}_t - (\sigma C / B)^{2}}, \gamma' \big)}.
\end{equation}

Equation \ref{eq:adam_corr} enables a sign descent interpretation for DP-Adam which closely tracks that of Adam. Ignoring the stochasticity introduced by measurements with DP noise for now, we have that:
\begin{enumerate}
    \item If for parameter $i$, $|\E{[\Bar{g}_t]}|_i \gg \sqrt{\V{[\Bar{g}_t]}_i + \Phi}$ , then $|\E{[\Bar{g}_t]}|_i \approx \sqrt{\E{[\Bar{g}^2_t]}_i}$, and $\Delta_t \approx \pm 1$. The update would be similar even without of our bias correction.
    \item If for parameter $i$, $|\E{[\Bar{g}_t]}|_i \gg \sqrt{\V{[\Bar{g}_t]}_i}$ but $|\E{[\Bar{g}_t]}|_i \ll \Phi$, then correcting for $\Phi$ ensures that $|\E{[\Bar{g}_t]}|_i \approx \sqrt{\E{[\Bar{g}^2_t]}_i}$, and $\Delta_t \approx \pm 1$, the expected behavior under Adam and the sign descent hypothesis. Without the correction, the update would be scaled as $\E{[\bar{g}_t]}/\Phi$ instead, and proportional to the gradient size, which is not the Adam or sign descent behavior.
    \item If for parameter $i$, $|\E{[\Bar{g}_t]}|_i \not\gg \sqrt{\V{[\Bar{g}_t]}_i}$ (large gradient variance), $\Delta_t \in [-1, 1]$, performing a smooth (variance scaled) version of sign descent (not correcting for $\Phi$ would make the update closer to $0$, especially if $\Phi$ is large compared to $\V{[\Bar{g}_t]}_i$).
\end{enumerate}
Of course, the exponential moving averages over DP noise introduce measurement errors: it is possible that $\hat{v}_{i,t} - \Phi < \V{[\bar{g}_t]}_i$ and even that $\hat{v}_{i,t} - \Phi <0$. Our stability correction, $\textrm{max}( . , \gamma')$, deals with these cases similarly to Adam's $\gamma$, and we expect that $\sqrt{\gamma'} \gg \gamma$. While it can still happen that 
$|\Delta_{i, t}| \geq 1$, we show in \S\ref{sec:exp} that debiasing the second moment to follow the sign descent interpretation yields an important improvement in model accuracy. Algorithm \ref{alg:dp_adam} shows the modified DP-Adam with a corrected estimate for the second moments.

\begin{algorithm}[htb]
\SetAlgoLined
\footnotesize
\KwOut{Model parameters $\theta$}
\KwIn{Data $D=\{x_i\}_{i=1}^N$, loss function $\gL$, $\eta$, $\sigma$, $B$, $C$, $\beta_1, \beta_2$, $\gamma'$, $\epsilon$-DP, $\delta$-DP; initialize $\theta_0$ randomly; $m_0 = 0, v_0 = 0$; total number of steps $T = f(\epsilon\textnormal{-DP}, \delta\textnormal{-DP}, B, N, \sigma)$}
 \For{$t = 1 \ldots, T$} {
    Take a random batch with sampling probability $B/N$ \;
    $g_n = \nabla \gL(\theta_{t-1}, x_n)$, $\forall x_n$ in the batch \;
    $\Tilde{g}_t = \frac{1}{B} \big( \sum_{i} g_{i} / {\max{\big(1, \frac{\lVert g_i \rVert_{2}}{ C}} \big)} + z_t \big), \; z_t \sim \cN\big(0, \sigma^{2}C^{2}\I^{d}\big)$ \;
    $m_{t} \leftarrow \beta_{1} \cdot m_{t-1} + \left(1-\beta_{1}\right) \cdot \Tilde{g}_t$,\, $\widehat{m}_{t} \leftarrow m_{t} /\left(1-\beta_{1}^{t}\right)$ \;
    $v_{t} \leftarrow \beta_{2} \cdot v_{t-1}+\left(1-\beta_{2}\right) \cdot \Tilde{g}_t^{2}$,\, $\widehat{v}_{t} \leftarrow v_{t} /\left(1-\beta_{2}^{t}\right)$ \;
    $\theta_{t} \leftarrow \theta_{t-1} - \eta \cdot \hat{m}_t / \sqrt{\max{\big(\hat{v}_t - (\sigma C / B)^{2}}, \gamma' \big)}$
    }
 \caption{DP-Adam (with corrected DP bias in second moment estimation)}
 \label{alg:dp_adam}
\end{algorithm}

\paragraph{Privacy Analysis.} 
Our bias corrected version of DP-Adam follows the same DP analysis as that of Adam without correction, and that of DP-SGD. Since both $\hat{m}_t$ and $\hat{v}_t$ are computed from the privatized gradient $\Tilde{g_t}$, the post-processing property of DP, and composition over training iterations, ensures privacy. The correction is based only on public parameters of DP-Adam: $\beta_{2}$, step $t$, batch size $B$, and the DP noise variance $(\sigma C)^{2}$. In experiments (\S\ref{sec:exp}) we use R\'enyi DP for composition, though other techniques would also apply.

\section{The empirical effect of Correcting for DP noise}
\label{sec:exp}
\begin{figure}[htb]
\centering
\includegraphics[width=0.9\linewidth]{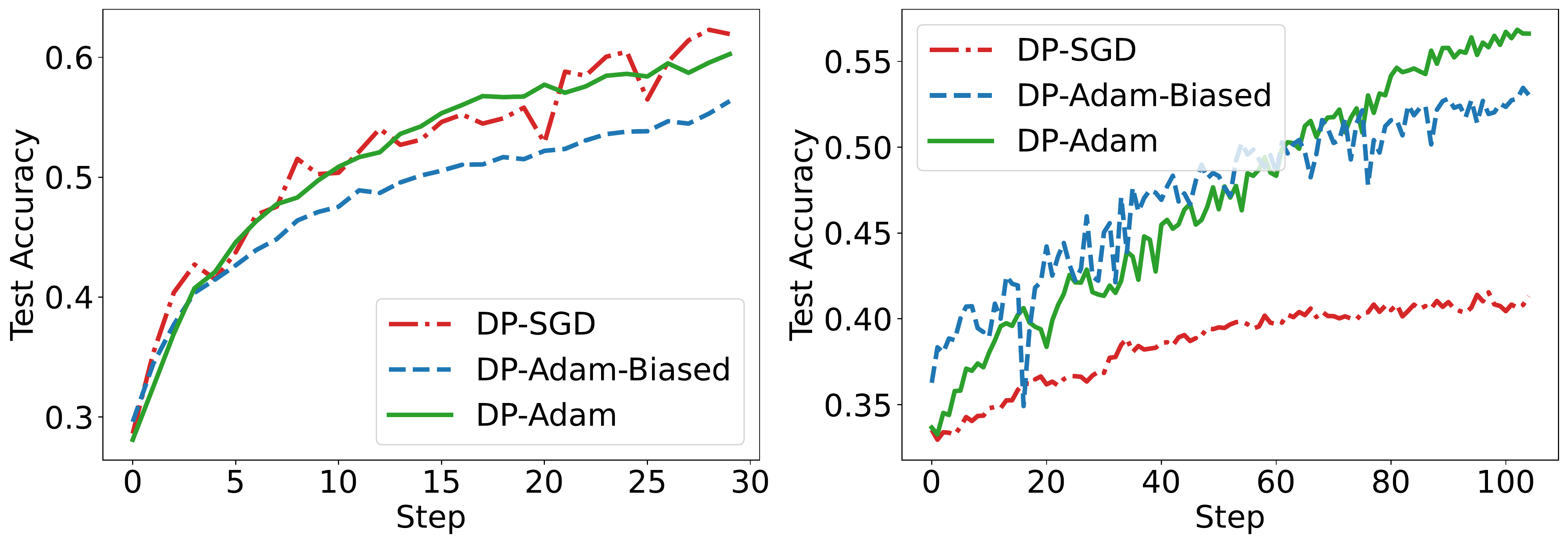}
\caption{Comparing the performance of DP-Adam, DP-Adam-Biased and DP-SGD on \textbf{Left:} CIFAR10 (images) and \textbf{Right:} SNLI (NLP). At the end of training $\epsilon\textnormal{-DP} \approx 7$. Each optimizer is tuned separately.}
\label{fig:exp1}
\end{figure}

\paragraph{Performance of DP-Adam, DP-Adam-Biased and DP-SGD.}
We compare the performance of DP-Adam, DP-Adam-Biased (with no correction in $v_t^{p}$) and DP-SGD on image, text and graph node classification tasks with CIFAR10 \citep{cifar10_data} and SNLI \citep{bowman-etal-2015-large}. We evaluate the training-from-scratch setting: for image classification, we use a 5-layer CNN model and all of the model parameters are initialized randomly; for text classification, only the last encoder and the classifier blocks are initialized randomly and the other layers inherit weights from pre-trained BERT-base model \citep{bert_paper}. For each optimizer, we tune the learning rate, as well as $\gamma$ or $\gamma'$, at a coarse granularity to maximize test accuracy at $\epsilon \approx 7$ for CIFAR10 and SNLI.
The final hyperparameters are: for CIFAR10, $C=1, \sigma=1, B=2048$, $\eta$, $\gamma$ (or $\gamma'$) for DP-SGD/DP-Adam-Biased/DP-Adam are respectively 4/0.001/0.001, (not applicable)/1e-8/1e-8; for SNLI, $C=0.1, \sigma=0.4, B=256$, $\eta$, $\gamma$ (or $\gamma'$) for DP-SGD/DP-Adam-Biased/DP-Adam are respectively 4/0.01/0.001, (not applicable)/1e-8/1e-9.
Figure \ref{fig:exp1} shows that for CIFAR10, on which Adam often performs worse than SGD in non-private settings, DP-Adam-Biased performs much worse than DP-SGD (56.35\% vs 61.94\% accuracy), whereas DP-Adam brings the performance close to that of DP-SGD (60.27\% accuracy, a $4$ percentage points improvement on DP-Adam-Biased). On SNLI, DP-Adam performs better than DP-Adam-Biased: the accuracy improves from 53.04\% to 56.62\% ($3.5$ percentage points). Both perform much better than DP-SGD (41.31\% on SNLI).

\paragraph{First and second moment estimates of clipped and private gradients.}
We numerically compare the scale of $v_t^{c}$ and $v_t^{p}$ by measuring their summary statistics at the early ($t=5000$) and late ($t=20000$) training stages. The results are summarized in Table \ref{tab:exp2}. We observe that both the scale and spread of $v_t^{p}$ is quite different from that of $v_t^{c}$, which suggests that the values of $v_t^{p}$ are largely affected by the DP noise.
If no correction is imposed, $\Phi$ dominates the size of $\V{[\overline{g}_t]}$, and the tuned learning rate is larger to compensate.
Since $\Phi$ dominates, the update $\Delta_t$ is proportional to the first moment $\E[\bar{g}_t]$ (\S\ref{sec:method}). This is not compatible with the behavior of sign descent.
\begin{table}[]
\centering
\resizebox{0.75\textwidth}{!}{%
\begin{tabular}{cccccccc}
\hline
\textbf{} &  & \textbf{Min} & \textbf{Q1} & \textbf{Median} & \textbf{Q3} & \textbf{Max} & \textbf{Mean} \\ \hline
\multirow{2}{*}{\textbf{t = 5000}} & $v_t^{c}$ & 4.757e-21 & 1.802e-13 & 6.333e-13 & 1.481e-12 & 2.647e-8 & 4.050e-12 \\
 & $v_t^{p}$ & 2.025e-8 & 2.363e-8 & 2.426e-8 & 2.463e-8 & 5.324e-8 & 2.436e-8 \\ \hline
\multirow{2}{*}{\textbf{t = 20000}} & $v_t^{c}$ & 6.584e-22 & 5.867e-14 & 2.925e-13 & 8.372e-13 & 1.060e-8 & 3.673e-12 \\
 & $v_t^{p}$ & 2.065e-8 & 2.408e-8 & 2.460e-8 & 2.513e-8 & 3.657e-8 & 2.461e-8 \\ \hline
\end{tabular}%
}
\caption{Summary statistics of $v_t^{p}$ with the SNLI dataset.}
\label{tab:exp2}
\end{table}

\begin{figure}[htb]
\centering
\includegraphics[width=\linewidth]{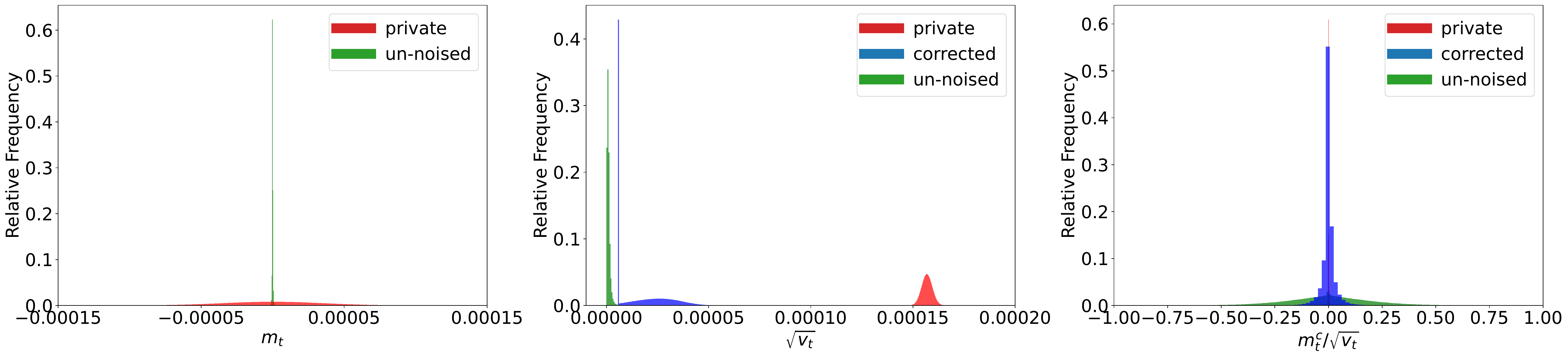}
\caption{Histogram of \textbf{Left: }clean ($m_t^c$) and biased ($m_t^p$) first moment estimates, \textbf{Middle: }clean ($v_t^c$), biased ($v_t^p$) and corrected ($v_t^p - \Phi$) second moment estimates, \textbf{Right: }clean ($m_t^{c}/\sqrt{v_t^c}$), biased ($m_t^{c}/\sqrt{v_t^p}$) and corrected ($m_t^{c}/\sqrt{v_t^p - \Phi}$) Adam updates with respect to $m_t^{c}$.
}
\label{fig:mt_vt_hist}
\end{figure}

To further study the effect of DP noise and of our bias correction, we compare the distribution of the private, un-noised, and corrected variables. We use the SNLI dataset for demonstration with $B=256, C=0.1, \sigma=0.4, \beta_2=0.999$, $\Phi \approx \textnormal{2.441e-8}$. Figure \ref{fig:mt_vt_hist} (Left) shows the histogram of un-noised ($m_t^c$) and private ($m_t^p$) first moment estimates.
We observe that the center of the distribution aligns, confirming that $\E{[m_t^p]} = \E{[m_t^c]}$ as in Equation \ref{eq:mt}. The private first moment distribution has larger variance compared to the clean distribution as a result of DP noise. Figure \ref{fig:mt_vt_hist} (Middle) shows the histogram of un-noised ($v_t^c$), private ($v_t^p$), and corrected ($v_t^p - \Phi$) second moment estimates. We see that the distributions of $v_t^c$ and $v_t^p$ are quite different, with a shift in the center approximately equal to $\sqrt{\Phi}$. This suggests that the DP noise variance dominates the scale of $v_t^p$ in Equation \ref{eq:vt}. The corrected second moment estimates are much closer in scale to the clean estimates, with the gap near 0 due to the effect of the numerical stability constant $\gamma'$. Figure \ref{fig:mt_vt_hist} (Right) shows the distribution of the un-noised ($m_t^{c}/\sqrt{v_t^c}$), private ($m_t^{c}/\sqrt{v_t^p}$) and corrected ($m_t^{c}/\sqrt{v_t^p - \Phi}$) Adam updates with respect to $m_t^{c}$. We observe that the un-noised distribution is mostly in $[-1, 1]$ whereas the private distribution is heavily concentrated around 0. The bias correction alleviates the concentration around 0 in the distribution, which is consistent with the interpretation in \S \ref{sec:method}. 

\begin{figure}[htb]
\centering
\includegraphics[width=0.85\linewidth]{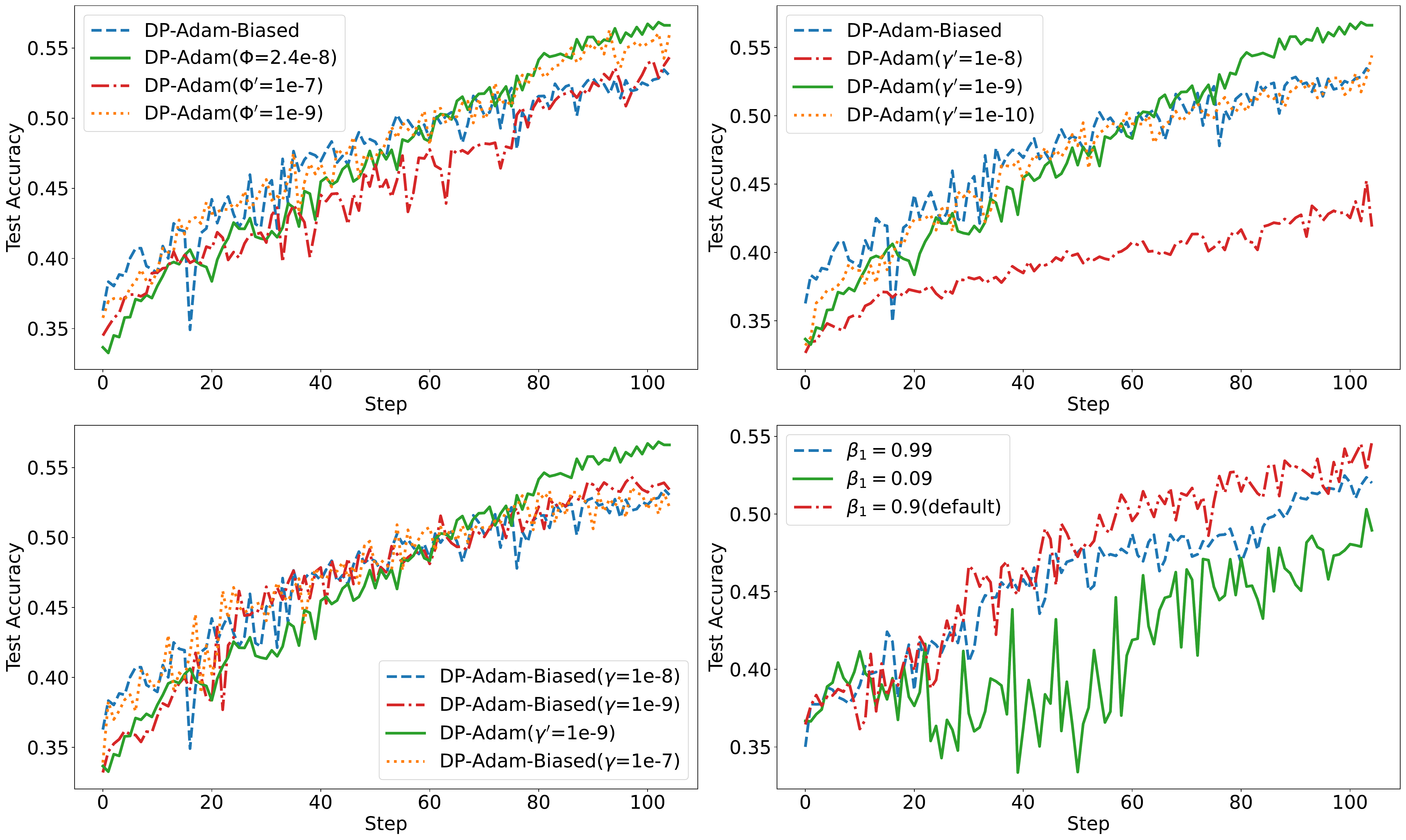}
\caption{Compare the performance when \textbf{Upper Left: }subtracting different (fake) values of $\Phi$,
\textbf{Upper Right: }tuning $\gamma'$ in DP-Adam, \textbf{Lower Left: }tuning $\gamma$ in DP-Adam-Biased, \textbf{Lower Right: }tuning $\beta$s in DP-Adam-Biased.
}
\label{fig:exp3_4}
\end{figure}

\paragraph{Correcting second moment with different values.}
We test whether the noise variance $\Phi$ is indeed the correct value to subtract from the noisily estimated $v_t^{p}$, by subtracting other values $\Phi'$ at different scales instead. In Figure \ref{fig:exp3_4}(Upper Left) we compare the performance of correcting $v_t^{p}$ with the true $\Phi$=2.4e-8 versus $\Phi'$. The experiments of DP-Adam($\Phi'$=1e-7) and DP-Adam($\Phi'$=1e-9) are trained using the same DP hyperparameters except changing value of $\Phi$ to $\Phi'$ and with coarsely tuned learning rates. We observe that both values of $\Phi'>\Phi$ or $\Phi'<\Phi$ lead to weaker performance. It suggests that the DP noise bias in the second moment estimate may be responsible for the degraded performance, and correcting for a different value does not provide a good estimate for $\E{[\overline{g}^2_t]}$.

\paragraph{Effect of the numerical stability constant.}
The numerical stability constant $\gamma$ in known to affect the performance of Adam in the non-private setting, and $\gamma$ is often tuned as a hyperparameter \citep{Reddi2019}. Following the same logic, we test the effect of $\gamma'$ and $\gamma$ on the performance of DP-Adam and DP-Adam-Biased. Figure \ref{fig:exp3_4} (Upper Right) shows that $\gamma'$ indeed impacts the performance of DP-Adam: values of $v_t^{p}$ are small, and changing $\gamma'$ can avoid magnifying a large number of parameters with tiny estimates of $v_t^{c}$. 
Figure \ref{fig:exp3_4} (Lower Left) shows the effect of tuning $\gamma$ in DP-Adam-Biased. We observe that it has a smaller effect than with DP-Adam, since the large scale of $\Phi$ makes the estimates of $v_t^{p}$ relatively large and similar among parameters. We also observe that tuning $\gamma$ with DP-Adam-Biased does not lead to the same effect as correcting $\Phi$ in DP-Adam, and DP-Adam achieves higher accuracy. 

\paragraph{Effect of the moving average coefficients.}
The $\beta$ coefficients control the effective length of the moving average window in Adam's estimates of the first and second moments. It thus balances the effect of averaging out the noise, versus estimating moments with older gradients. A larger $\beta$ implies averaging over a longer sequence of past gradients, which potentially benefits performance by decreasing the effect of noise. Figure \ref{fig:exp3_4} (Lower Right) shows the effect of choosing different $\beta$ in DP-Adam-Biased, with the learning rate $\eta$ coarsely tuned from 1e-4 to 1e-2. As suggested in \citet{orig_adam}, we set $\beta_1$ and choose $\beta_2$ such that $(1-\beta_1) = \sqrt{1-\beta_2}$. We observe that setting $\beta$s too large or too small is worse than choosing the default values ($\beta_1=0.9, \beta_2=0.99$). Setting $\beta$ smaller shows a clear disadvantage as the performance is both worse and more volatile due to less smoothing over noise. Setting a larger $\beta$ results in similar performance at the end of training. However, lowering the effect of noise this way does not yield similar improvements as correcting for DP noise bias in the second moments.

\bibliography{ref}
\bibliographystyle{iclr2021_conference}

\end{document}